# Automated Radiographic Total Sharp Score (ARTSS) in Rheumatoid Arthritis: A Solution to Reduce Inter-Intra Reader Variation and Enhancing Clinical Practice

1   **Hajar Moradmand, Ph.D.,** Department of Radiation Oncology, University of Maryland School of Medicine, Baltimore, MD, USA
2   **Lei Ren, Ph.D.,** Department of Radiation Oncology, University of Maryland School of Medicine, Baltimore, MD, USA

**Short title**: ARTSS in Rheumatoid Arthritis

*Correspondence: Hajar Moradmand, 1 Department of Radiation Oncology, University of Maryland School of Medicine, Baltimore, MD, USA
Tel: 202-340-3730

Email Address: hmoradmand@som.umaryland.edu

## Data availability
The hand X-rays utilized in this study are publicly available and were obtained from previous research).

The dataset can be downloaded from the following link:

https://drive.google.com/drive/folders/1O11ROJypqyBksrfYqTIjdXxdc6MBao27?usp=sharing

## Conflict of Interest:
The authors declare no conflict of interest and no financial support for this work that could have influenced its outcome.

## Author contribution
All authors contributed to interpreting the data, writing, and editing the paper. H.M. undertook to model and draft the paper, supervised by L.R. conceived the study with support from all co-authors.

## Ethics approval

There has been no need for ethical approval from an institutional review board or ethics committee before commencing this study.

## Abstract

Assessing the severity of rheumatoid arthritis (RA) using the Total Sharp/Van Der Heijde Score (TSS) is crucial, but manual scoring is often time-consuming and subjective.

This study introduces an Automated Radiographic Sharp Scoring (ARTSS) framework that leverages deep learning to analyze full-hand X-ray images, aiming to reduce inter- and intra-observer variability. The research uniquely accommodates patients with joint disappearance and variable-length image sequences. We developed ARTSS using data from 970 patients, structured into four stages: I) Image pre-processing and re-orientation using ResNet50, II) Hand segmentation using UNet.3, III) Joint identification using YOLOv7, and IV) TSS prediction using models such as VGG16, VGG19, ResNet50, DenseNet201, EfficientNetB0, and Vision Transformer (ViT).

We evaluated model performance with Intersection over Union (IoU), Mean Average Precision (MAP), mean absolute error (MAE), Root Mean Squared Error (RMSE), and Huber loss.

The average TSS from two radiologists was used as the ground truth. Model training employed 3-fold cross-validation, with each fold consisting of 452 training and 227 validation samples, and external testing included 291 unseen subjects.

Our joint identification model achieved 99% accuracy. The best-performing model, ViT, achieved a notably low Huber loss of 0.87 for TSS prediction.

Our results demonstrate the potential of deep learning to automate RA scoring, which can significantly enhance clinical practice. Our approach addresses the challenge of joint disappearance and variable joint numbers, offers timesaving benefits, reduces inter- and intra-reader variability, improves radiologist accuracy, and aids rheumatologists in making more informed decisions.

*Keywords*: Automated Radiographic Sharp Scoring, Deep Learning, Joint Identification, Rheumatoid Arthritis, X-ray Images.

## Introduction

Rheumatoid arthritis (RA) is a chronic autoimmune disease characterized by inflammation that predominantly targets small joints but can advance to larger joints and other organs, leading to significant complications like joint deformity, bone erosion, and weakening of tendons and ligaments [1]. In the United States, RA affects roughly 1.5 million individuals, equivalent to 0.6% of the adult population, while globally, the prevalence approximates 1% [2].

These statistics emphasize the substantial impact of RA on both national and international levels, highlighting the significance of developing innovative methodologies for its detection and evaluation. The complications of RA encompass not only its direct impact on the confines of joints but also its potential to extend its effects to larger joints, such as the knees and hips, which can become involved as the disease progresses. Furthermore, systemic embodiments may appear, leading to complications affecting organs like the heart, lungs, and blood vessels. Thus, the consequences of RA span a diverse range of clinical presentations, warranting a comprehensive approach to its diagnosis and management.

Recent advances in cellular profiling techniques like single-cell transcriptomics and spatial transcriptomics have led to the discovery of novel pathogenic cell types in RA joint tissues, highlighting the significant heterogeneity in cellular composition among RA patients. These discoveries provide exciting opportunities for precision medicine approaches to RA treatment[3]. Precision medicine in RA involves using genomic profiling, biomarker testing, and imaging techniques to customize treatment plans for each patient[4]. Joint damage is a hallmark of RA that can lead to deformities and loss of function, and radiographic imaging is an essential tool for disease assessment[5]. The total Sharp/van der Heijde score (TSS) is a widely used clinical scoring method that can objectively quantify joint damage and aid in tracking disease progression [6]. This method evaluates erosion and joint space narrowing (JSN) in 16 joints of each hand and six joints of each foot. Erosion and JSN are rated on a scale from 0 to 5 and 0 to 4, respectively. Although the TSS is a valuable tool in precision medicine approaches to RA treatment scoring systems, it is time-consuming and subject to inter- and intra-reader variability, potentially limiting its clinical effectiveness [7, 8]. Computerized analysis, like deep convolutional neural networks (CNNs), has gradually become prevalent for RA [9] leading to increased interest in their application for automated TSS and the launch of the RA2-DREAM challenge [10].

Although there is growing research interest in automated joint identification and TSS assessment [11, 12], it has not yet been widely adopted in clinical settings due to challenges in achieving reproducible and generalized workflows and results across different radiographic images. These challenges arise from variations in image quality, differences in imaging equipment and protocols at various medical centers, and changes in imaging methods, such as including both hands in a single frame or using bilateral views of hands. Additionally, technical issues like improper angling of the image receiver or axis further complicate the process.

Identifying joints in X-ray images with deformative hands and severe erosion, where joints may disappear, results in a variable number of joints for each patient, making it challenging for computerized object recognition models, like the difficulties faced by human observers. Consequently, these cases are often excluded from previous research, limiting the generalizability and reproducibility of these models for clinical use.

To address these issues, we developed a standardized ARTSS framework that considers the most plausible variables affecting the replicability of our findings. We trained and tested our models on publicly available data and employed custom data augmentation techniques to balance the dataset, particularly for cases with TSS > 100. Additionally, we devised a novel approach to handle cases with varying numbers of joints, especially in patients with severe joint erosion, ensuring the inclusion of relevant patient data often excluded in prior studies. Finally, we established a reliable pipeline to enhance the generalizability of our findings.

The study utilized advanced deep learning architectures, specifically U-net and YOLOv7[13], for hand segmentation and joint identification. We also employed several well-established pre-trained models, including VGG16 [14], VGG19, EfficientNet [15], ResNet [16], and DenseNet [17], as the backbone for predicting radiographic scores from patient data, leveraging their extensive knowledge acquired through large datasets and complex pre-training objectives. Moreover, we developed a custom Vision Transformer (ViT) [18] model for the same purpose. Cross-validation, data augmentation, and the utilization of Huber loss were integral methods employed to optimize the performance and evaluation metrics of the models. An overview of the workflow for analyzing hand X-ray images is presented in Figure1.

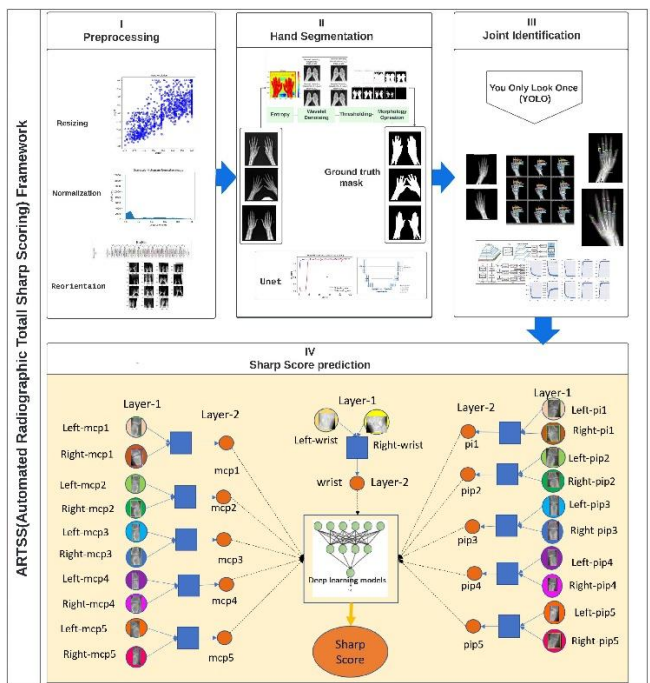

Figure 1. Overview of the workflow. The ARTSS (automated radiographic total Sharp scoring) framework consists of four stages: Image-preprocessing and orientation, Hand Masking, joint identification, and Sharp score prediction.

The ARTSS system shows promise in reducing inter- and intra-reader variability in RA X-ray evaluation, thereby enhancing the quality of care for patients. With the widespread availability of digital radiographs, implementing automatic scoring systems like ARTSS can facilitate the assessment of joint damage in RA patients, especially in centers with limited resources where expert readers may not be easily accessible.

## METHODS

### Study population

The dataset utilized in this study comprises 970 hand X-ray images from previously published literature with publicly available data[19]. These hand X-ray images were acquired from patients who underwent treatment at the Department of Rheumatology at the Clinical Medical College of Anhui University of Traditional Chinese Medicine between 2019 and 2021. The dataset consists of 803 female and 167 male patients, with an average age of 52 years. Figure 2 shows the distribution of age and Total Sharp/Van Der Heijde Score (TSS) of patients across genders. The distribution reveals a slightly higher frequency of lower scores with a peak of around 40-50 years for both genders. Each subplot displays a histogram overlaid with kernel density estimation (KDE) curves.

Two experienced radiologists scored the images independently according to the Total Sharp/van der Heijde method. The average scores were utilized for subsequent evaluation to consider inter- and intra-rater reliability and consistency.

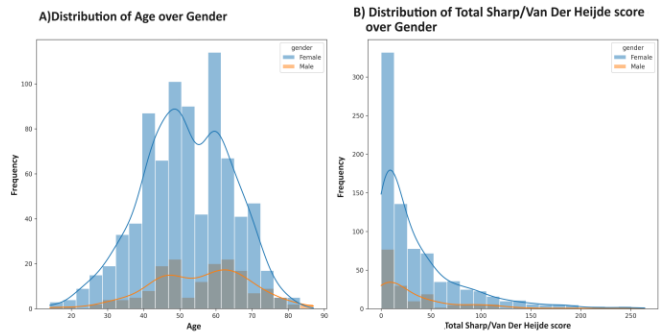

Figure 2. Distribution of Age and Sharp Score over Gender. A) Histogram of age by gender with KED curves shows a peak of around 40-50 and 60-70 years among female and male patients. B) Histogram of Total Sharp Score over gender shows the imbalance of scores with a higher frequency of lower scores for both genders. Kernel density estimation (KDE) curves provide a smooth representation of the distributions, with histograms indicating the frequency distribution of age and Total Sharp Score values.

### Image Pre-Processing

We performed several preprocessing steps on our dataset, which included image resizing, normalization, and reorientation. Image resizing was necessary to ensure uniform input sizes for our model, which is essential for effective processing. Normalization was applied to standardize pixel values across all images and make easier comparisons and consistency of outcomes.

Additionally, we aligned all images to a 90-degree orientation using ResNet50. This step was crucial for improving the generalizability of our models, as hand X-ray images can vary in orientation due to differences in imaging equipment and patient positioning.

## Hand Segmentation

We utilized the U-Net architecture, a widely used method for semantic hand segmentation. To create accurate ground truth masks, we developed a four-step process:

1. Applied a Gaussian filter to the images to smooth them out and reduce noise.

2. Used wavelet transformation to denoise the image's entropy.

3. Applied a threshold to the image to segment the hand region.

4. Refined the segmented masks using morphological operations.

Out of the initial dataset of 970 images, our algorithm processed 582 images to generate masks. We manually selected 382 ground truth masks based on visual evaluation.

## Joint Identification

We utilized the YOLOv7 for joint identification, an improved version of the popular YOLO (You Only Look Once) [20] object detection algorithm. Specifically, we targeted the proximal interphalangeal (PI), proximal interphalangeal (PIP), metacarpophalangeal (MCP), and wrist as region of interest (ROI). Initially, we evaluated our model on 428 training, 112 testing, and 430 validation sets. However, due to the small amount of data and the complexity of joint identification, the model failed to distinguish between left and right hands appropriately. To tackle these limitations and enhance the model's performance, we cropped the left and right hands, simplified joint localization, and augmented the dataset to 856 training images, 224 validation images, and 860 testing images. The annotation of 11880 joints (PI, PIP-1, PIP_2, PIP_3, PIP_4, MCP_0, MCP_1, MCP_2, MCP_3, MCP_4, and wrist for each hand) was manually performed on 1080 images using MakeSense.ai, an open-source web-based toolkit.

## Data Augmentation

We employed image data augmentation [21] techniques to address the problem of limited data, enhancing its diversity and boosting our deep learning models 's capacity to generalize to novel data instances. The augmentation process included rescaling the pixel values to a normalized range of 0 to 1 (rescale=1.0 / 255.0), random rotations up to 10 degrees (rotation range=10), and width and height shifts of up to 20% of the image dimensions (width_shift_range=0.2, height_shift_range=0.2). Additionally, horizontal flipping (horizontal_flip=True) was applied, along with adjustments to brightness levels (brightness_range= [0.7, 1.2]).

## Handling Variable-Length Image Sequences

Differences in the visible number of joints among patients, attributed to factors such as hand deformities and joint disappearance in moderate to severe cases, present a challenge in standardizing data for deep-learning models. To tackle this issue, we devised a preprocessing technique.

Initially, we identified the maximum number of joint images across all patients in our dataset. Subsequently, we implemented a padding approach for sequences with fewer images, aligning them with this maximum length. This methodology ensured uniform sequence length across all inputs, thus facilitating consistent data processing within our models.

During model training, we employed masking techniques to prevent the padded regions of the input sequences from influencing the model's predictions. This strategy ensured that the model focused solely on the relevant information within each sequence. As a result, our approach maintained the length uniformity of all input sequences, thereby enabling consistent feeding into the models.

## Statical analysis

In this investigation, we assessed the performance of hand segmentation and joint identification models with IoU (Intersection over Union) and MAP (Mean Average Precision) metrics. Additionally, we utilized the Huber loss function to address the imbalance of the Sharp scoring range. These four-evaluation metrics are described in detail below.

I. Intersection over Union (IoU): IoU is a widely used metric in image segmentation tasks [22]. IoU is calculated as the ratio of the area of intersection between the predicted and ground truth masks ($|A \cap B|$) to the area of their union ($|A \cup B|$), as indicated in (1). Higher IoU values indicate better alignment between the predicted and ground truth masks.

$$IoU = \frac{|A \cap B|}{|A \cup B|} \quad (1)$$

II. *Mean Average Precision(MAP):* MAP is a widely used metric in object detection to evaluate how accurately the model detects objects across different categories [23]. It calculates the average precision (AP) for each object category by examining the precision-recall curve. The average precision represents the area under this curve. MAP then computes the average of these average precision values across all categories. This provides a comprehensive evaluation of the model's ability to detect objects effectively. MAP ranges from 0 to 1, where higher values indicate better performance. The formula for MAP is represented by (2), considering parameters like the number of ground truth objects ($N_{pos}$), the ranked list of predictions ($R_k$), $|classes|$ the total number of classes, and $AP_c$ the average precision for class c.

$$AP = \frac{1}{N_{pos}} \sum_{k=1}^{N_{pos}} Precision(R_k) \quad (2)$$

$$MAP = \frac{1}{|classes|} \sum_{c \in classes} AP_c$$

III. *Huber Loss Function:* Huber loss [24], also known as the Smooth Mean Absolute Error, is frequently used in regression problems, particularly when dealing with outliers or noisy data. It presents a robust alternative to traditional loss functions like the mean squared error (MSE) and mean absolute error (MAE). It behaves like MSE for smaller deviations and MAE for large errors. Equation (3) represents the formula for the Huber loss function.

$$L_\sigma(y, f(x)) = \begin{cases} \frac{1}{2}(y - f(x))^2 & |y - f(x)| \leq \sigma \\ \sigma|y - f(x)| - \frac{1}{2}\sigma^2 & |y - f(x)| > \sigma \end{cases} \quad (3)$$

Where $L_\sigma$ denotes the Huber loss function, y represents the actual target value, f(x) denotes the predicted value by the regression model, and δ is a threshold parameter. In simpler terms, it calculates the loss differently depending on how big the error is, using a δ.

## RESULTS

After training a U-Net model on a dataset of hand images, we achieved precise segmentation of the hand region from the background in each radiographic image. The trained U-Net model demonstrated high accuracy, with a mean Intersection over Union (IoU) score of 0.94 on a held-out test set. In Figure 3 (A) the loss and IOU score of U-Net for hand segmentation on training and validation is presented. The graph shows that as the number of epochs increases, the training and validation loss decreases, and after 20 epochs it remains relatively constant.

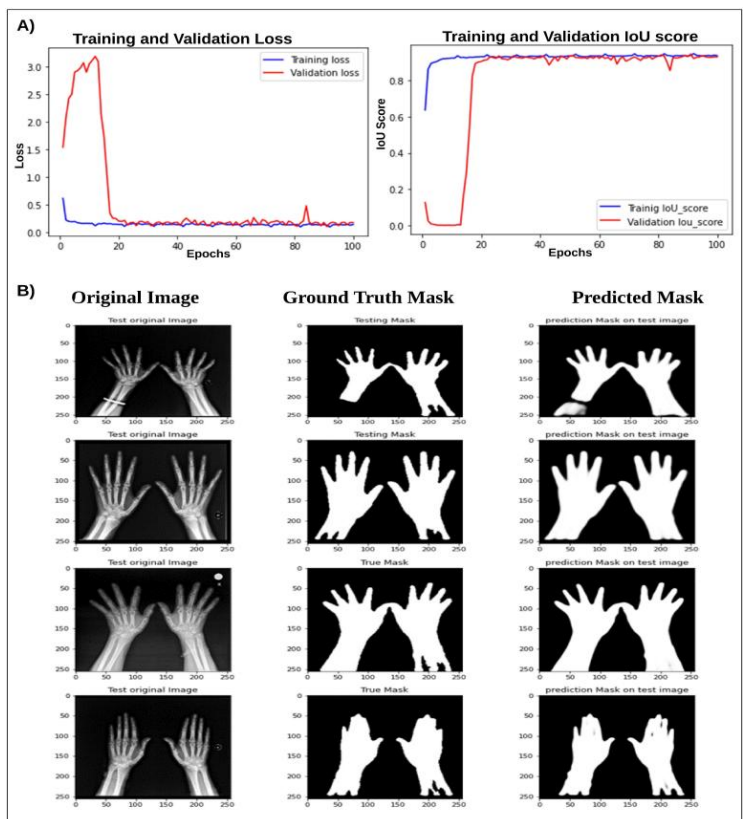

Figure 3. Hand segmentations using U-Net. A) U-Net Training and Validation Loss and IoU (Intersection over union) score. The graph shows that as the number of epochs increases, the training and validation loss decreases, and after 20 epochs it remains relatively constant. Similarly, the IoU score for both training and validation sets increases with each epoch, indicating that the model is improving over time. B) Sample of prediction of mask using U-Net on four hand X-ray images.

Similarly, the IoU score for both training and validation sets increases with each epoch, indicating that the model is improving over time. Figure 4(A) presents the performance analysis and sample results of joint identification using YOLOv7.

In addition, we evaluated the effectiveness of different deep learning models in predicting TSS using metrics such as mean absolute error (MAE), root mean squared error (RMSE), and Huber loss. Model training included a 3-fold cross-validation approach, each fold containing 452 training and 227 validation samples. An external test was performed on 291 unseen subjects to validate the generalizability of the models. Table 1 summarizes the results for each model on the test dataset.

Table 1. Performance of Different Models on rheumatoid arthritis Hand Radiographs for Total Sharo scoring (TSS) prediction on the test dataset.

| Model | Loss | Fold 1 | Fold 2 | Fold 3 | Average |
|---|---|---|---|---|---|
| VGG16 | MAE | 3.98 | 3.38 | 3.20 | 3.52 |
|  | RMSE | 4.5 | 4.00 | 3.90 | 4.13 |
|  | Huber Loss | 3.64 | 3.10 | 3.01 | 3.25 |
| VGG19 | MAE | 3.02 | 2.96 | 2.98 | 2.99 |
|  | RMSE | 3.50 | 3.45 | 3.40 | 3.45 |
|  | Huber Loss | 3.14 | 2.93 | 2.84 | 2.97 |
| EfficientNetB0 | MAE | 9.00 | 7.95 | 8.26 | 8.40 |
|  | RMSE | 9.8 | 8.60 | 8.91 | 9.10 |
|  | Huber Loss | 8.70 | 7.71 | 8.11 | 8.17 |
| ResNet50 | MAE | 7.95 | 5.9258 | 5.025 | 6.30 |
|  | RMSE | 8.20 | 6.5 | 7.61 | 7.43 |
|  | Huber Loss | 7.81 | 5.83 | 5.10 | 6.24 |
| DenseNet201 | MAE | 7.61 | 7.53 | 7.65 | 7.59 |
|  | RMSE | 8.13 | 8.38 | 6.42 | 7.64 |
|  | Huber Loss | 6.43 | 6.38 | 6.42 | 6.41 |
| Vision Transformer | MAE | 0.99 | 0.92 | 0.94 | 0.95 |
|  | RMSE | 0.89 | 0.94 | 0.98 | 0.93 |
|  | Huber Loss | 0.85 | 0.87 | 0.89 | 0.87 |

MAE: Mean Absolute Error, RMSE: Root Mean Squared Error

## DISCUSSION

Our Automated Radiographic Total Sharp Score (ARTSS) system reduces inter- and intra-reader variation by leveraging advanced deep learning models for consistent, objective joint assessment. Our method standardizes image analysis, eliminating the subjective biases common in manual scoring. Employing data augmentation and robust training strategies, the model reliably handles diverse radiographic images. This automation enhances reproducibility and streamlines clinical workflows, enabling accurate, efficient evaluation of joint damage.

Our study demonstrates a significant advancement in the automatic segmentation and joint identification in RA hand X-rays. By utilizing the ResNet50 architecture for orientation and a combination of U-Net and YOLOv7 for segmentation and joint identification, we achieved an accuracy of 99%. This surpasses the 87% accuracy achieved by Radke KL, et al.[25] accuracy using RetinaNet, and aligns closely with Chaturvedi N, who reported a near-perfect identification rate using RetinaNet, as presented in Table 2.

Table 2. Comparison segmentation and identification of Region of Interest (ROI) in X-ray images of rheumatoid arthritis between current and previous studies.

| Study | ROI | Orientation | Segmentation/Joint Identification | Result (Accuracy) |
|---|---|---|---|---|
| This Study | PI, PIP, MCP, and wrist \Hand | ResNet50 | U-Net / YOLOv7 | 99% |
| Radke KL, et al[25] | PI, PIP, MCP, and wrist \Hand | NA | RetinaNet | 87% |
| Dimitrovsky I [26] | PI, PIP, MCP, wrist\ Hand and foot | ResNet50 | Manually using mini-GUI | NA |
| Honda S[27] | PI, PIP, MCP, and wrist\Hand | ResNet34 | Heat map regression based on U-Net | NA, all were within 10 pixels of the correct coordinates |
| Tan YM, et al [28] | PI, PIP, MCP \ Hand | NA | U-Net / YOLOv3 | 94.5% |
| Wang HJ, et al[29] | PIP, MCP, and wrist \Hand | NA | Yolov4 | 92% |
| Üreten et al. [30] | Hand (RA, OA, and normal) | NA | Yolov4 | 90.7% |
| Fung DLX, et al.[31] | PIP, MCP, Wrist\Hand | NA | YOLOv516 | 0.99 |

In comparison to the previous study that used 226 hand X-ray images of 40 rheumatoid arthritis patients and DeepLabCut for detection and four classification models for intact/non-intact joint classification, our study utilized a larger dataset and a reliable pipeline to ensure robust and reproducible results in our ARTSS system.

Our original dataset comprises full images of both right and left hands. For joint identification, we trained the YOLOv7 model separately on images of individual left and right hands. Each hand contains a maximum of 11 identifiable regions of interest (ROI) joints.

As illustrated in Fig. 4(C), this model effectively identifies joints with varying numbers of joints. Specifically, Fig. 4(C.I.) shows the model successfully detecting all joints. However, Fig. 4(C.II) presents instances where the model misses some joints due to hand deformation, and Fig. 4(C.III) depicts severe cases where some joints disappear.

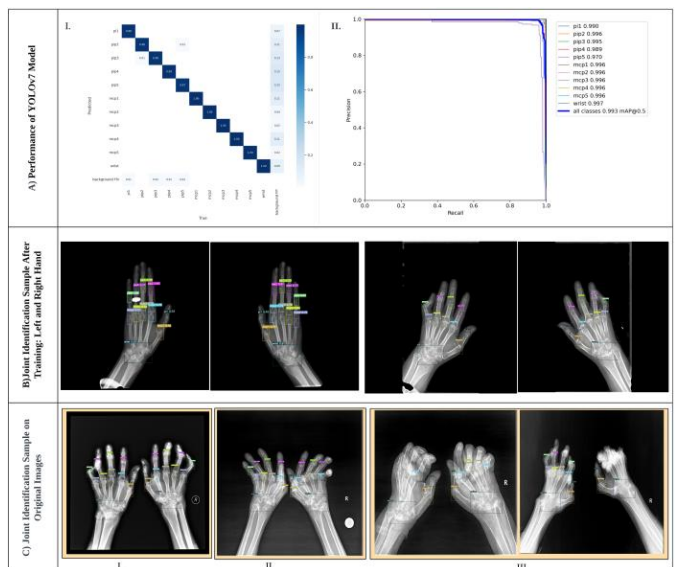

Figure 4. Performance Analysis and Sample Results of Joint Identification Using YOLOv7. A) The performance metrics of YOLOv7, a confusion matrix (A. I), and precision-recall curves (A.II). B) Samples of joint identification after training, focusing on left-hand and right-hand joint identification. C) Example of joint identification on 90-degree re-oriented images, including where the model successfully identifies all joints (C. I), some joints are missed due to deformative hands (C.II) and joint disappear (C.III).

Thus, the number of joints identified in each patient may vary in practical applications of the Automated Radiographic Total Sharp Scoring (ARTSS) system. Nevertheless, the input dimensions for the deep learning model must be the same.

To our knowledge, this study is the first research that addresses and provides a practical solution for handling variations in the number of joint images across patients by employing padding and masking techniques to ensure the same input lengths for the model. The results (Table I) indicate that the ViT model outperforms the other models across all three metrics including MAE, RMSE, and Huber Loss. VGG19, and VGG16 also show strong performance, closely following the Vision Transformer. EfficientNet, DenseNet, and ResNet show relatively higher error values in comparison.

Although our findings are robust, this work has some limitations. First, our dataset was obtained from a single center, which may limit the generalizability of our findings. Second, our 90-reorient model is constrained to images with orientations between 0 and 180 degrees, which may not encompass the full range of potential radiographic angles. Third, our study was limited to hand radiographs, leaving it unclear whether the ARTSS system would perform equally well on the feet joint.

Despite these limitations, our method remains an important step forward in clinical automated TSS systems for RA patients.

## CONCLUSION

In conclusion, our study provides a robust pipeline that considers factors potentially affecting the reproducibility of our finding. The ability of the ARTSS to include patients with a sever condition (higher TSS), which excluded from pervious studies, with different number of joints identified is a big step forward for improve generalizability and clinical applicability of this system.

The ARTSS system promises to improve RA patient care by reducing inter-reader variability in hand X-rays evaluation and decreasing the workload and time required for experienced readers to evaluate large volumes of radiographic images, especially in clinical with limited access to experienced physicians. The ARTSS framework has significant implications for disease monitoring, treatment decision making, and predicting disease outcomes.

For future research, we recommend further assessment of the pipeline developed in this study to confirm its clinical applicability. Additionally, we strongly encourage the establishment of a public multicenter database for RA patients that includes comprehensive information such as blood tests, molecular markers, and medical imaging data. This resource will facilitate additional studies and enable more valuable and comparable analyses by researchers globally. Integrating multidimensional data (pan-omics) and high-throughput image data (radiomics) [32, 33] with clinical assessments, such as disease activity and response to treatment, may further enhance the accuracy and utility of automatic scoring systems for RA assessment, may further enhance the accuracy and utility of automatic scoring systems for RA assessment.

## ACKNOWLEDGMENT

The authors would like to acknowledge the contributions of Chuanfu Li et al. from the Institute of Hefei Comprehensive National Science Center for providing a publicly accessible data source utilized in this study.

## SUPPORTING INFORMATION